\pdfoutput=1

\documentclass[11pt]{article}

\usepackage{naacl2021}

\usepackage{times}
\usepackage{latexsym}
\usepackage{graphicx}

\usepackage{booktabs}

\usepackage[T1]{fontenc}

\usepackage[utf8]{inputenc}

\usepackage{microtype}

\title{Semi-supervised Interactive Intent Labeling}

\author{Saurav Sahay \qquad Eda Okur \qquad Nagib Hakim \qquad Lama Nachman \\
Intel Labs, USA \\
\texttt{\{saurav.sahay,eda.okur,nagib.hakim,lama.nachman\}@intel.com} \\
}

\begin{document}
\maketitle
\begin{abstract}
Building the Natural Language Understanding (NLU) modules of task-oriented Spoken Dialogue Systems (SDS) involves a definition of intents and entities, collection of task-relevant data, annotating the data with intents and entities, and then repeating the same process over and over again for adding any functionality/enhancement to the SDS. In this work, we showcase an Intent Bulk Labeling system where SDS developers can interactively label and augment training data from unlabeled utterance corpora using advanced clustering and visual labeling methods. We extend the Deep Aligned Clustering~\cite{zhang2021discover} work with a better backbone BERT model, explore techniques to select the seed data for labeling, and develop a data balancing method using an oversampling technique that utilizes paraphrasing models. We also look at the effect of data augmentation on the clustering process. Our results show that we can achieve over 10\% gain in clustering accuracy on some datasets using the combination of the above techniques. Finally, we extract utterance embeddings from the clustering model and plot the data to interactively bulk label the samples, reducing the time and effort for data labeling of the whole dataset significantly.
\end{abstract}

\section{Introduction}

Acquiring an accurately labeled corpus is necessary for training machine learning (ML) models in various classification applications. Labeling is an expensive and labor-intensive activity requiring annotators to understand the domain well and to label the instances one at a time. In this work, we explore the task of labeling multiple intents visually with the help of a semi-supervised clustering algorithm. The clustering algorithm helps learn an embedding representation of the training data that is well-suited for downstream labeling. In order to label, we further reduce the high dimensional representation using the UMAP~\cite{2018arXivUMAP}. Since utterances are short, uncovering their semantic meaning to group them together is very challenging. SBERT~\cite{reimers-2019-sentence-bert} showed that out-of-the-box BERT~\cite{DBLP:journals/corr/abs-1810-04805} maps sentences to a vector space that is not very suitable to be used with common measures like cosine-similarity and euclidean distances. This happens because in the BERT network, there is no independent sentence embedding computation, which makes it difficult to derive sentence embeddings. Researchers utilize the mean pooling of word embeddings as an approximate measure of the sentence embedding. However, results show that this practice yields inappropriate sentence embeddings that are often worse than averaging GloVe embeddings~\cite{pennington-etal-2014-glove, reimers-2019-sentence-bert}. Many researchers have developed sentence embedding methods: Skip-Thought~\cite{NIPS2015_f442d33f}, InferSent~\cite{conneau-etal-2017-supervised}, USE~\cite{DBLP:journals/corr/abs-1803-11175}, SBERT~\cite{reimers-2019-sentence-bert}. State-of-the-art SBERT adds a pooling operation to the output of BERT to derive a fixed-sized sentence embedding and fine-tunes a Siamese network on the sentence-pairs from the NLI~\cite{bowman-etal-2015-large, DBLP:journals/corr/WilliamsNB17} and STSb~\cite{DBLP:journals/corr/abs-1708-00055} datasets. 

The Deep Aligned Clustering (DAC)~\cite{zhang2021discover} introduced an effective method for clustering and discovering new intents. DAC transfers the prior knowledge of a limited number of known intents and incorporates a technique to align cluster centroids in successive training epochs. The limited known intents are used to pre-train the model. The authors use the pre-trained BERT model~\cite{DBLP:journals/corr/abs-1810-04805} to extract deep intent features, then pre-train the model with a randomly selected subset of labeled data. The pre-trained parameters are used to obtain well-initialized intent representations. K-Means clustering is performed on the extracted intent features along with a method to estimate the number of clusters and the alignment strategy to obtain the final cluster assignments. The K-Means algorithm selects cluster centroids that minimize the Euclidean distance within the cluster. Due to this Euclidean distance optimization, clustering using the SBERT model to extract feature embeddings naturally outperforms other embedding methods. In our work, we have extended the DAC algorithm with the SBERT as an embedding backbone for clustering of utterances.

In semi-supervised learning, the seed set is selected using a sampling strategy: ``A simple random sample of size $n$ consists of $n$ individuals from the population chosen such that every set of $n$ individuals has an equal chance to be the sample actually selected.''~\cite{moore1989introduction}. However, these sample subsets may not represent the original data adequately because randomization methods do not exploit the correlations in the original population. In a stratified random sample, the population is classified first into groups (called strata) with similar characteristics. Then a simple random sample is chosen from each strata separately. These simple random samples are combined to form the overall sample. Stratified sampling can help ensure that there are enough observations within each strata to make meaningful inferences. DAC uses the Random Sampling method for seed selection. In this work, we have explored a couple of stratified sampling approaches for seed selection in hope to mitigate the limitations of random sampling and improve the clustering outcome.  

Another issue we address in this work is class sample imbalance. Seed selection generally yields an imbalanced dataset, which in turn impairs the predictive capability of the classification algorithms~\cite{Douzas_2018}. Some methods manipulate the training data, aiming to change the class distribution towards a more balanced one by undersampling or oversampling~\cite{kotsiantis2006handling, galar2011review}. SMOTE~\cite{chawla2002smote} is a popular oversampling technique proposed to improve random oversampling. In one variant of SMOTE, borderline minority instances are heuristically selected and linearly interpolated to create synthetic samples. In this work, we take inspiration from the SMOTE method and choose borderline minority instances and paraphrase them using a Sequence to Sequence Paraphrasing model. The paraphrases provide natural and meaningful augmentations of the dataset that are not synthetic. 

Previous work has shown that data augmentation can boost performance on text classification tasks~\cite{barzilay-mckeown-2001-extracting, dolan-brockett-2005-automatically, DBLP:journals/corr/abs-1708-00391, hu-etal-2019-large}.~\citet{wieting-etal-2017-learning} used Neural Machine Translation (NMT)~\cite{DBLP:journals/corr/SutskeverVL14} to translate the non-English side of the parallel text to get English-English paraphrase pairs. This method has been scaled to generate large paraphrase corpora~\cite{wieting-gimpel-2018-paranmt}. Prior work in learning paraphrases has used autoencoders~\cite{10.5555/2986459.2986549}, encoder-decoder architectures as in BART~\cite{DBLP:journals/corr/abs-1910-13461}, and other learning frameworks such as NMT~\cite{sokolov2020neural}. Data augmentation using paraphrasing is a simple yet effective strategy that we explored in this work to improve the clustering. 

\begin{figure*}[!t]
\caption{Interactive Labeling System Architecture}
\label{architecture}
\begin{center}
\includegraphics[width=\textwidth]{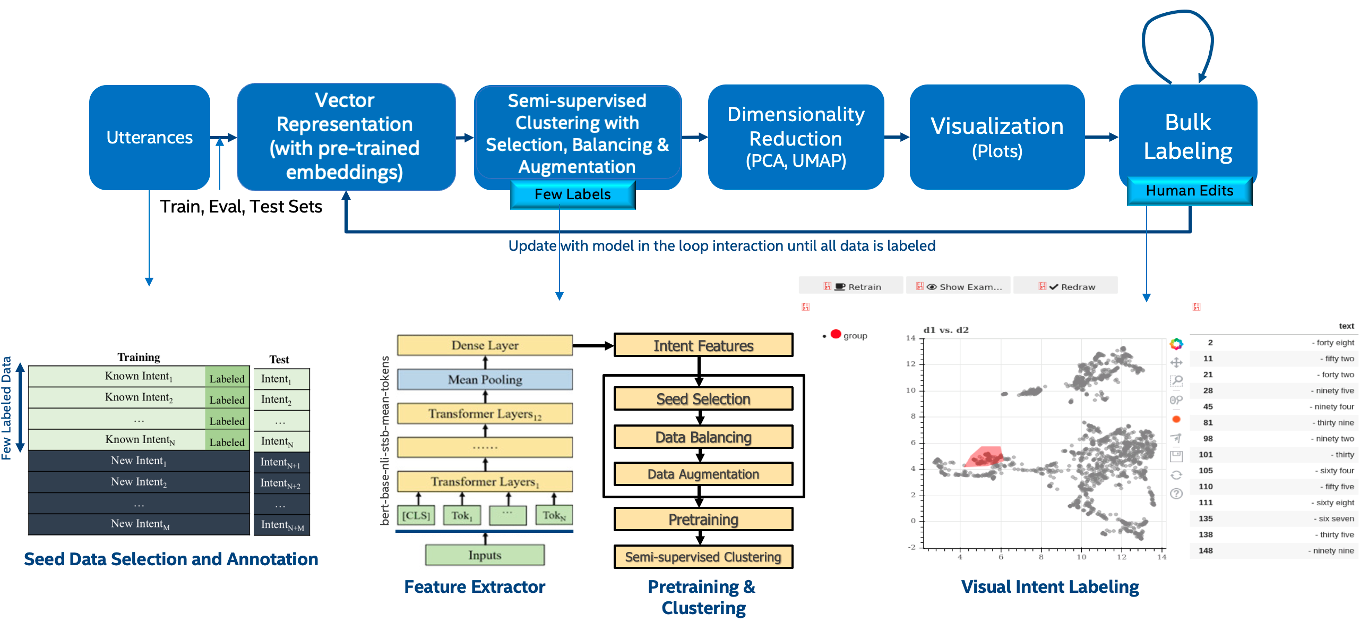}
\end{center}
\end{figure*}

For interactive visual labeling of utterances, we build up from the learnt embedding representation of the data and fine-tune it using the clustering. DAC learns to cluster with a weak self-supervised signal to update its representation and to optimize both local (via K-Means) and global information (via cluster alignment). This results in an optimized intent-level feature representation. This high dimensional latent representation can be reduced to 2-3 dimensions using the Uniform Manifold Approximation and Projection (UMAP)~\cite{2018arXivUMAP}. We use Rasa WhatLies\footnote{\url{rasahq.github.io/whatlies/}} library~\cite{warmerdam-etal-2020-going} to extract the UMAP embeddings. For interactive labeling, we utilize an interactive visualization library called Human Learn\footnote{\url{koaning.github.io/human-learn/}}~\cite{vincent_d_warmerdam_2021_4708001} that allows us to draw decision boundaries on a plot. By building on top of the work of Rasa Bulk Labelling\footnote{\url{github.com/RasaHQ/rasalit/blob/main/notebooks/bulk-labelling/bulk-labelling-ui.ipynb}} UI~\cite{youtube, bokeh}, we augment the interface with our learnt representation for interactive labeling. Although we focus on NLU, other studies like `Conversation Learner'~\cite{shukla-etal-2020-conversation} focus on interactive dialogue managers (DM) with human-in-the-loop annotations of dialogue data via machine teaching. Note also that although the majority of task-oriented SDS still involves defining intents/entities, there are recent examples that argue for a richer target representation than the classical intent/entity model, such as SMCalFlow~\cite{10.1162/tacl_a_00333}.

\section{Methodology}
Figure~\ref{architecture} describes the semi-supervised labeling process. We start with the unlabeled utterance corpus and apply seed sampling methods to select a small subset of the corpus. Once the selected subset is manually labeled, we address the data imbalance with our paraphrase-based minority oversampling method. We can also augment the labeled corpus with paraphrasing to provide more data for the clustering process. The DAC algorithm is applied with improved embeddings to extract the utterance representation for interactive labeling.

\begin{table*}[!h]
  \centering
  \small
  \begin{tabular}{lcccccc}
    \toprule
    Dataset & \#Classes & \#Train & \#Valid & \#Test & Vocab & Length (max / mean) \\
    \midrule
    CLINC & 150 & 18,000 & 2,250 & 2,250 & 7,283 & 28 / 8.31\\
    BANKING & 77 & 9,003 & 1,000 & 3,080 & 5,028 & 79 / 11.91\\
    KidSpace & 19 & 1,289 & 445 & 419 & 2,581 & 74 / 5.10\\
    \bottomrule
  \end{tabular}
  \caption{Dataset Statistics}
  \label{data-st}
\end{table*}

\subsection{Sentence Representation}
For sentence representation, we use the HuggingFace Transformers model BERT-base-nli-stsb-mean-tokens\footnote{https://huggingface.co/sentence-transformers/bert-base-nli-stsb-mean-tokens/tree/main}. This model was first fine-tuned on a combination of Stanford Natural Language Inference (SNLI)~\cite{bowman-etal-2015-large} (570K sentence-pairs with labels contradiction, entailment, and neutral) and Multi-Genre Natural Language Inference~\cite{DBLP:journals/corr/WilliamsNB17} (430K diverse sentence-pairs with same labels as SNLI) datasets, then on Semantic Textual Similarity benchmark (STSb)~\cite{DBLP:journals/corr/abs-1708-00055} (provide labels between 0 and 5 on the semantic relatedness of sentence pairs) training set. This model achieves a performance of 85.14 (Spearman’s rank correlation between the cosine-similarity of the sentence embeddings and the gold labels) on STSb regression evaluation. For context, the average BERT embeddings achieve a performance of 46.35 on this evaluation~\cite{reimers-2019-sentence-bert}.

\subsection{Seed Selection}
We explore two selection and sampling strategies for seed selection as follows:
\begin{itemize}
  \item \textbf{Cluster-based Selection (CB)}: In this method, we apply K-Means clustering on the $N$ utterances to partition the data into $n$ seed number of subsets. For example, if 10\% of the data has 100 utterances, this method creates 100 clusters from the dataset. We then pick the centroid's nearest neighbor as part of the seed set for all the clusters. The naive intuition for this strategy is that it would create a large number of clusters spread all over the data distribution ($N/n$ instances per cluster on average for uniformly distributed instances).    
  \item \textbf{Predicted Cluster Sampling (PCS)}: This is a stratified sampling method where we first predict the number of clusters and then sample instances from each cluster. We use the cluster size estimation method from the DAC work as follows: K-Means is performed with a large $K'$ (initialized with twice the ground truth number of classes). The assumption is that real clusters tend to be dense and the cluster mean size threshold is assumed to be N/K'. 
        \[K=\sum_{i=1}^{K'} \delta(|S_i|>=t) \]
where \(|S_{i}|\) is the size of the $i${th} produced cluster, and \(\delta(condition)\) is an indicator function. It outputs 1 if condition is satisfied, and outputs 0 if not. The method seems to perform well as reported in DAC work.
\end{itemize}

\subsection{Data Balancing and Augmentation}
For handling data imbalance, we propose a paraphrasing-based method to over-sample the minority classes. The method is described as follows:
\begin{enumerate}
 \item For every instance $p_i$ ($i = 1 , 2 , ... , p_{num}$) in the minority class $P$, we calculate its $m$ nearest neighbors from the whole training set $T$. The number of majority examples among the $m$ nearest neighbors is denoted by $m'$ ($0 \leq m' \leq m$). 
\item If $m' = m$ , i.e., all the $m$ nearest neighbors of $p_i$ are majority examples, $p_i$ is considered to be noise and is not operated in the following steps. If $\frac{m}{2} \leq m' < m$, namely the number of $p_i$’s majority nearest neighbors is larger than the number of its minority ones, $p_i$ is considered to be easily misclassified and put into a set DANGER. If $0 \leq m' < \frac{m}{2}$, $p_i$ is safe and does not need to participate in the following steps. 
\item The examples in DANGER are the borderline data of the minority class $P$, and we can see that DANGER $\subseteq P$. We set DANGER $= \{p'_1 , p'_2 , ... , p'_{d_{num}}\}$, $0 \leq d_{num} \leq p_{num}$
\item For each borderline data (that can be easily misclassified), we paraphrase the instance. For paraphrasing, we fine-tuned the BART Sequence to Sequence model~\cite{DBLP:journals/corr/abs-1910-13461} on a combination of 3 datasets: ParaNMT~\cite{wieting-gimpel-2018-paranmt}, PAWS~\cite{paws2019naacl, pawsx2019emnlp}, and the MSRP corpus~\cite{dolan-brockett-2005-automatically}.
\item We classify the paraphrased sample with a RoBERTa~\cite{DBLP:journals/corr/abs-1907-11692} based classifier fine-tuned on the labeled data and only add the instance if the classifier predicts the same label as the minority instance. We call this the `ParaMote' method in our experiments. Without this last step (5), we call this overall approach our `Paraphrasing' method.
\end{enumerate}
We use the Paraphrasing model and the classifier as a data augmentation method to augment the labeled training data (refer to as `Aug' in our experiments).

Note that we augment the paraphrased sample if it belongs to the same minority class (`ParaMote') as we do not want to inject noise while solving the data imbalance problem. The opposite is also possible for other purposes such as generating semantically similar adversaries~\cite{ribeiro-etal-2018-semantically}.


\section{Experimental Results}
To conduct our experiments, we use the BANKING~\cite{casanueva-etal-2020-efficient} and CLINC~\cite{larson-etal-2019-evaluation} datasets similar to the DAC work~\cite{zhang2021discover}. We also use another dataset called KidSpace that includes utterances from a Multimodal Learning Application for 5-to-8 years-old children~\cite{sahay-etal-2019-modeling, 10.1145/3279981.3279986}. We hope to utilize this system to label future utterances into relevant intents. Table~\ref{data-st} shows the statistics of the 3 datasets where 25\% random classes are kept unseen at pre-training. 

\begin{table*}[!t]
  \centering
  \footnotesize
  \begin{tabular}{lcccccc}
    \toprule
    Dataset & BERT & Data Bal/Aug & Seed Selection & NMI & ARI & ACC \\
    \midrule
    \textbf{BANKING} & Standard & None & RandomSampling & 79.22 & 52.96 & 63.84$\pm$1.91 \\
     & & & ClusterBased & 78.51 & 51.53 & 63.73$\pm$1.73 \\
     & & & PredictedClusterSampling & 78.62 & 51.72 & 62.72$\pm$0.97 \\
     \cmidrule(r){2-7}
      & Sentence & None & RandomSampling & 82.96 & 60.72 & 71.27$\pm$2.28 \\
     & & & ClusterBased & 80.65 & 55.03 & 65.44$\pm$1.24 \\
     & & & PredictedClusterSampling & 82.11 & 58.43 & 69.78$\pm$2.08 \\
     \cmidrule(r){2-7}
     & Sentence & Paraphrasing & RandomSampling &	83.00 & 60.95 & 71.95 \\
     & & & PredictedClusterSampling & 82.20 & 58.86 & 69.62 \\
     \cmidrule(r){2-7}
     & Sentence & ParaMote & RandomSampling & 82.58 & 59.54 & 69.92 \\
     & & & PredictedClusterSampling & 81.88 & 58.13 & 69.74 \\
     \cmidrule(r){2-7}
     & Sentence & Aug (3x) & RandomSampling & 82.94 & 60.78 & 71.66 \\
     & & & PredictedClusterSampling & 81.69 & 58.18 & 69.99 \\
    \midrule
    \textbf{CLINC} & Standard & None & RandomSampling & 93.90 & 79.70 & 86.34$\pm$1.47 \\
     & & & ClusterBased & 90.60 & 69.60 & 77.87$\pm$1.70 \\
     & & & PredictedClusterSampling & 93.76 & 79.42 & 86.41$\pm$0.65 \\
     \cmidrule(r){2-7}
     & Sentence & None & RandomSampling & 93.80 & 79.06 & 85.76$\pm$1.17 \\
     & & & ClusterBased & 90.25 & 67.23 & 74.25$\pm$1.83 \\
     & & & PredictedClusterSampling & 93.60 & 78.57 & 85.43$\pm$0.96 \\
     \cmidrule(r){2-7}
     & Sentence & Paraphrasing & RandomSampling &	93.78 & 79.14 & 85.86 \\
     & & & PredictedClusterSampling & 93.40 & 77.68 & 84.89 \\
     \cmidrule(r){2-7}
     & Sentence & ParaMote & RandomSampling & 93.79 & 79.10 & 85.81 \\
     & & & PredictedClusterSampling & 93.48 & 77.97 & 84.86 \\
     \cmidrule(r){2-7}
     & Sentence & Aug (3x) & RandomSampling & 93.69 & 78.67 & 85.52 \\
     & & & PredictedClusterSampling & 92.96 & 76.50 & 83.96 \\
    \midrule
    \textbf{KidSpace} & Standard & None & RandomSampling & 71.40 & 48.26 & 58.55$\pm$4.22 \\
     & & & ClusterBased & 68.13 & 39.26 & 53.48$\pm$4.47 \\
     & & & PredictedClusterSampling & 70.53 & 45.33 & 56.80$\pm$4.56 \\
     \cmidrule(r){2-7}
      & Sentence & None & RandomSampling & 75.62 & 63.41 & 68.66$\pm$4.96 \\
     & & & ClusterBased& 71.27 & 53.16 & 62.10$\pm$9.59 \\
     & & & PredictedClusterSampling& 75.74 & 61.99 & 67.04$\pm$7.66 \\
     \cmidrule(r){2-7}
      & Sentence & Paraphrasing & RandomSampling & 76.41 & 63.02 & 68.83 \\
     & & & PredictedClusterSampling & 75.52 & 61.53 & 68.21 \\
     \cmidrule(r){2-7}
      & Sentence & ParaMote & RandomSampling & 76.28 & 61.20 & 68.09 \\
     & & & PredictedClusterSampling & 76.33 & 62.05 & 68.21 \\
     \cmidrule(r){2-7}
     & Sentence & Aug (3x) & RandomSampling & 76.48 & 61.33 & 68.07 \\
     & & & PredictedClusterSampling & 76.37 & 58.97 & 68.78 \\
    \bottomrule
  \end{tabular}
  \caption{Semi-supervised DeepAlign Clustering Results with BERT Model, Data Balance/Augmentation and Seed Selection on BANKING, CLINC, and KidSpace datasets (averaged results over 10 runs with different seed values; labeled ratio is 0.1 for BANKING and CLINC, 0.2 for KidSpace; known class ratio is 0.75 in all cases)}
  \label{all-summary}
\end{table*}

\subsection{Sentence Representation}
The choice of pre-trained embeddings has the largest impact on the clustering results. We observe huge performance gains for the single domain KidSpace and BANKING datasets. For the multi-domain and diverse CLINC dataset with the largest number of intents, we saw a slight degradation in performance. While this needs further investigation, we believe the dataset is diverse enough and already has very high clustering scores and that the improved sentence representations may not be helping further.

\subsection{Seed Selection}
\begin{figure}[t]
\caption{Cluster Visualization}
\label{clusters_centroid}
\begin{center}
\includegraphics[width=0.5\textwidth]{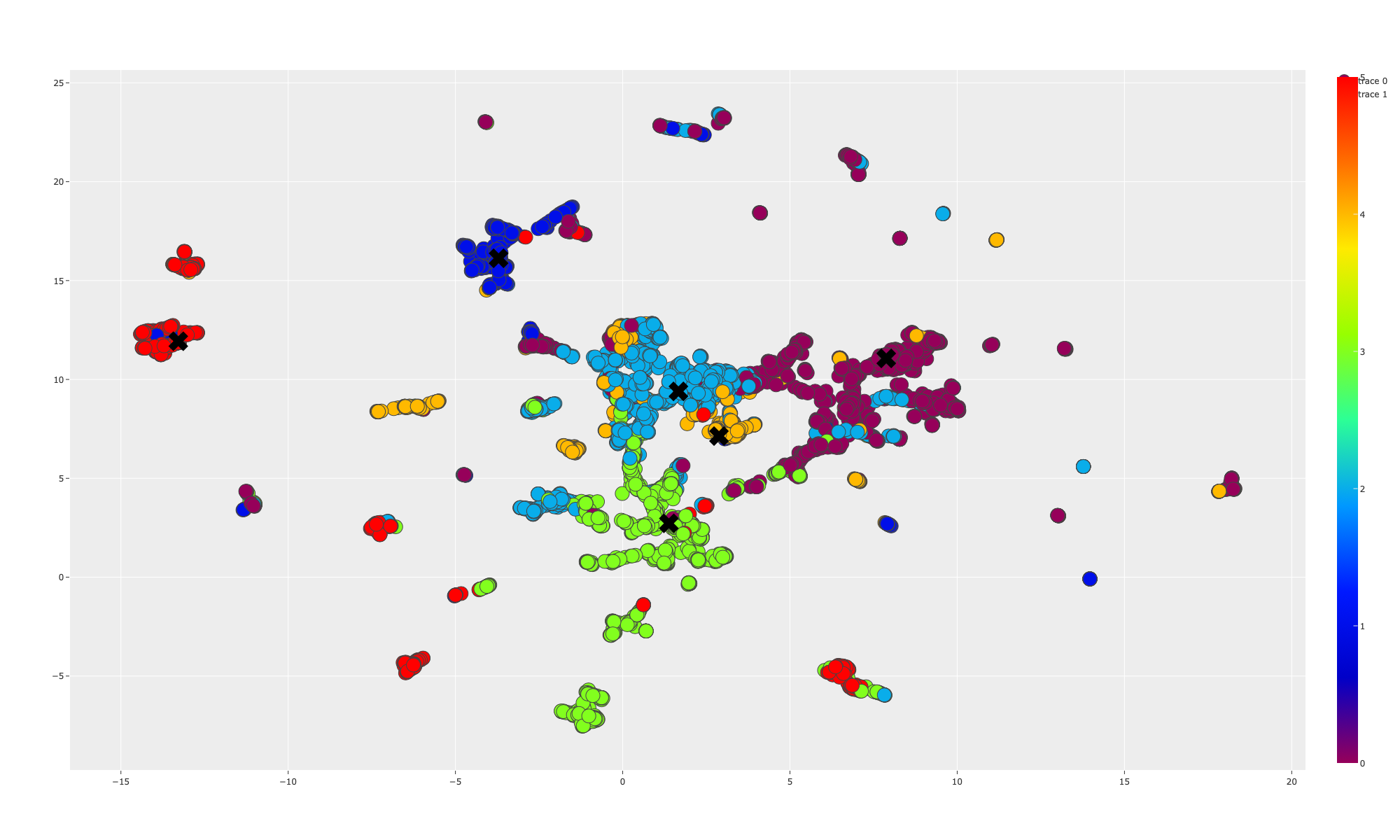}
\end{center}
\end{figure}

Seed selection is an important problem for limited data tasks. Law of large numbers does not hold and random sampling strategy may lead to larger variance in outcomes. We explored Cluster-based Selection (CB) and Predicted Cluster Sampling (PCS) besides other techniques (see detailed results in Appendix~\ref{appendix}). Our results trend towards smaller standard deviations and similar performance for the BANKING and CLINC datasets with the PCS method. Surprisingly, this does not hold for the KidSpace dataset that needs further investigation. Figure~\ref{clusters_centroid} shows the KidSpace data visualised with various colored clusters and centroids. While we non-randomly choose seed data, we still hide 25\% of the classes at random (to enable unknown intent discovery). Our recommendation is to use PCS if one cannot run the training multiple times for certain situations to have less variance in results.

\subsection{Data Balancing for Imbalanced Data}
\begin{figure}[t]
\caption{Label Distribution}
\label{histogram}
\centering
\includegraphics[width=0.5\textwidth]{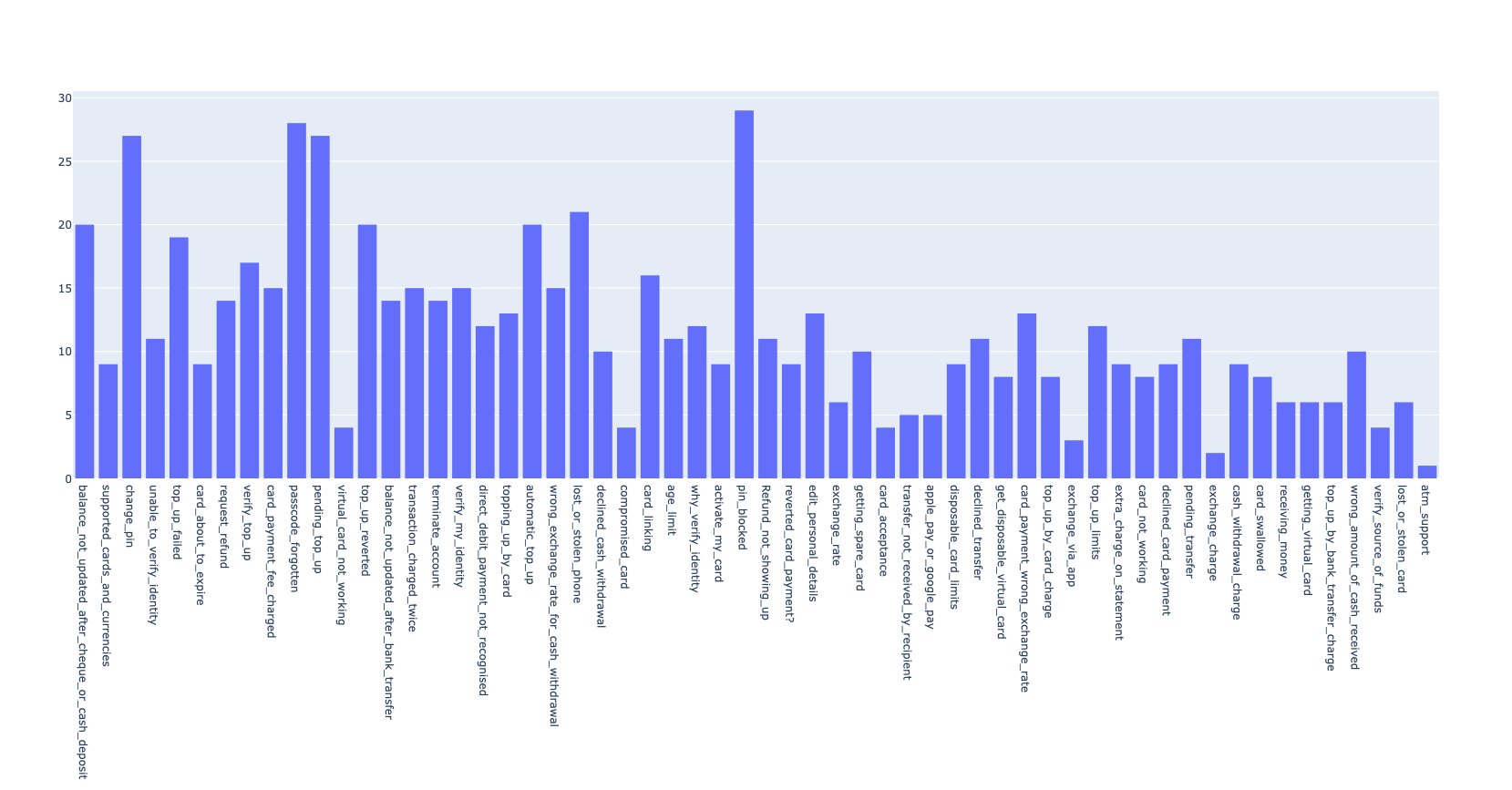}
\end{figure}

Figure~\ref{histogram} shows the histogram for the seed data, which is highly imbalanced and may adversely impact the clustering performance. We apply Paraphrasing and ParaMote methods to balance the data. Paraphrasing almost always improves the performance while the additional classifier to check for class-label consistency (ParaMote) does not help.  

\subsection{Data Augmentation}
We augmented the entire labeled data including the majority class using Paraphrasing (with class-label consistency) by 3x in our experiments. We aimed to understand if this could help get a better pre-trained model that could eventually improve the clustering outcome. We do not observe any performance gains with the augmentation process.  

\subsection{Interactive Data Labeling}
\begin{figure}[t]
\caption{Cluster Visualization on KidSpace with BERT-base/SBERT w/wo pre-training}
\label{clusters_bert_sbert_ks}
\begin{center}
\includegraphics[width=0.5\textwidth]{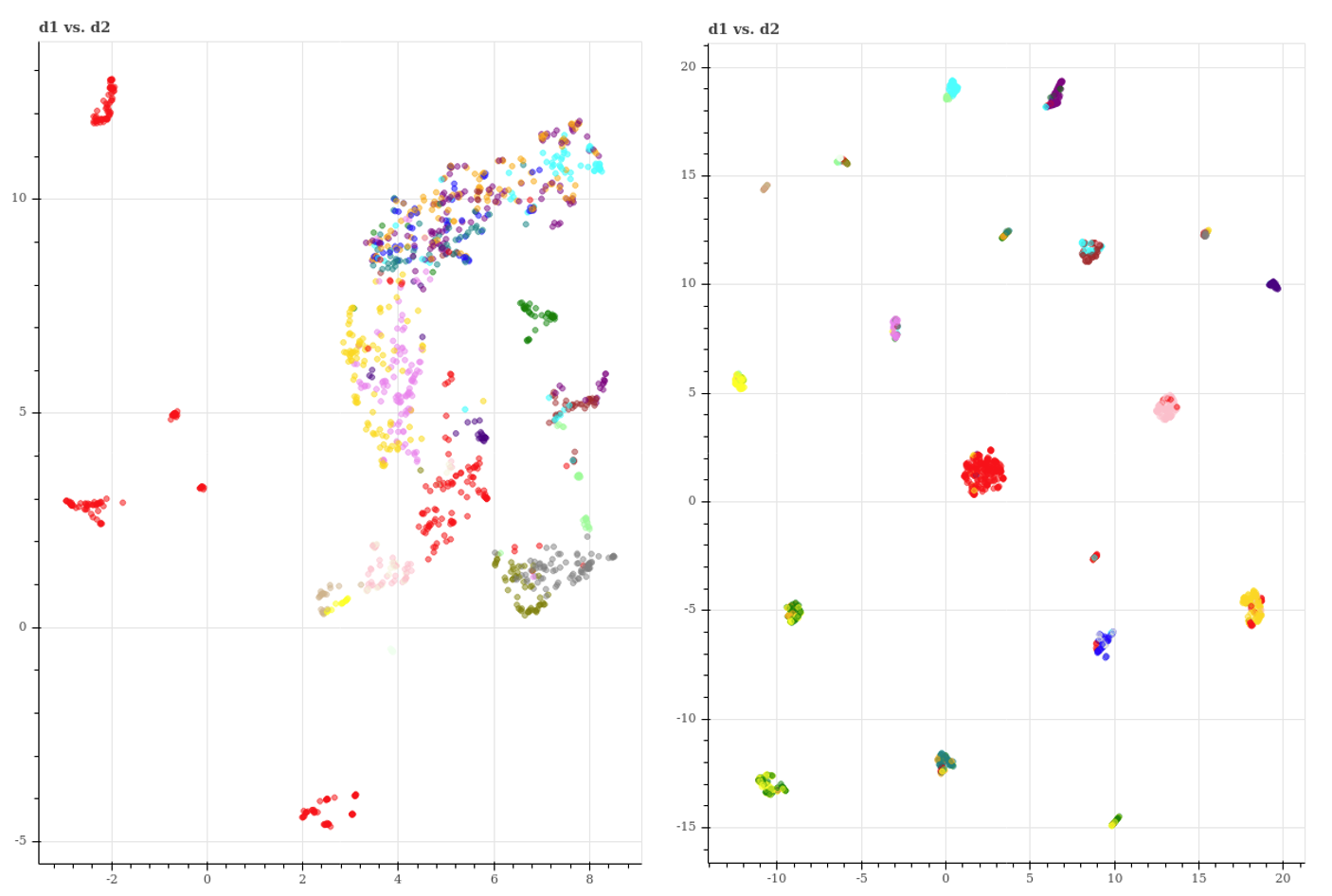}
\end{center}
\end{figure}

\begin{figure}[t]
\caption{Cluster Mixup on KidSpace due to Game Semantics}
\label{clusters_sbert_confusion}
\begin{center}
\includegraphics[width=0.5\textwidth]{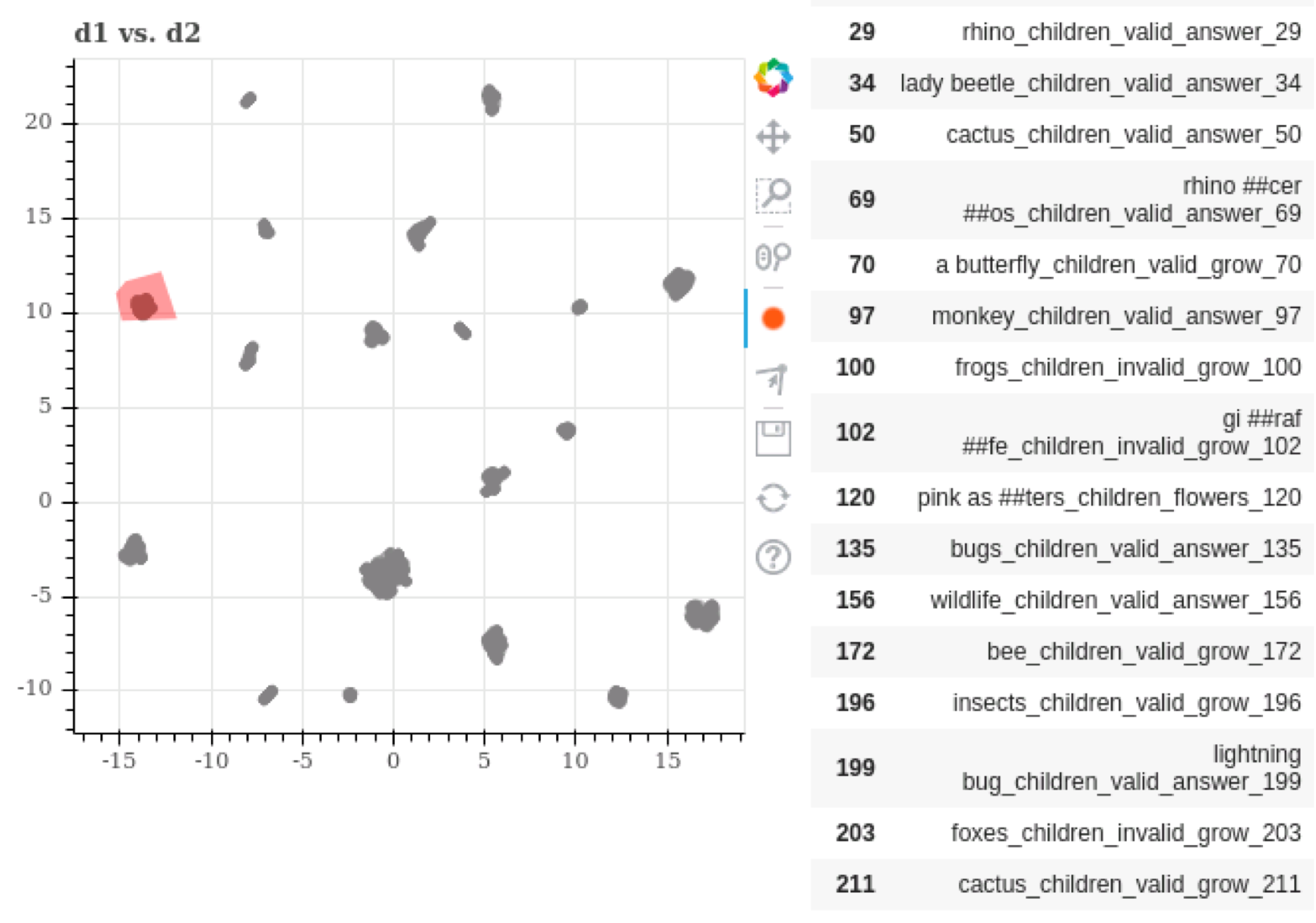}
\end{center}
\end{figure}

Our goal in this work is to develop a well-segmented learnt representation of the data with deep clustering and then to use the learnt representation to enable fast visual labeling. Figure~\ref{clusters_bert_sbert_ks} shows the two clustered representations, one without pre-training and BERT-base embedding while the other with a fine-tuned sentence BERT representation and pre-training. We can obtain well separated visual clusters using the latter approach. We use the drawing library human-learn to visually label the data. Figure~\ref{clusters_sbert_confusion} shows selected region of the data with various labels and class confusion. We notice that this representation not only helps with the labeling but also helps with correcting the labels and identify utterances that belong to multiple classes which cannot be easily segmented. For example, `children-valid-answer' and `children-invalid-grow' (invalid answers) contain semantically similar content depending on the game logic of the interaction. We perhaps need to group these together and use an alternative logic for implementing game semantics. 

\subsection{Conclusion}
In this exploration, we have used fine-tuned sentence BERT model to significantly improve the clustering performance. Predicted Cluster Sampling strategy for seed data selection seems to be a promising approach with possibly lower variance in clustering performance for smaller data labeling tasks. Paraphrasing-based data imbalance handling slightly improves the clustering performance as well. Finally, we have utilized the learnt representation to develop a visual intent labeling system.


\bibliography{anthology,custom}
\bibliographystyle{acl_natbib}

\begin{table*}[!t]
  \centering
  \footnotesize
  \begin{tabular}{lcccccc}
    \toprule
    Dataset & BERT & Seed Selection & labeled\_ratio & NMI & ARI & ACC \\
    \midrule
     \textbf{BANKING} & Standard & RandomSampling & 0.1 & 79.22 & 52.96 & 63.84$\pm$1.91 \\
     & & ClusterBased & 0.1 & 78.51 & 51.53 & 63.73$\pm$1.73 \\
     & & KnownClusterBased & 0.1 & 68.86 & 34.87 & 47.76$\pm$1.81 \\
     & & ClusterBasedSentenceEmb & 0.1 & 79.35 & 53.89 & 65.83$\pm$1.22 \\
     & & PredictedClusterSampling & 0.1 & 78.62 & 51.72 & 62.72$\pm$0.97 \\
     \cmidrule(r){2-7}
      & Sentence & RandomSampling & 0.1 & 82.96 & 60.72 & 71.27$\pm$2.28 \\
     & & ClusterBased & 0.1 & 80.65 & 55.03 & 65.44$\pm$1.24 \\
     & & KnownClusterBased & 0.1 & 74.21 & 42.37 & 53.89$\pm$2.53 \\
     & & ClusterBasedSentenceEmb & 0.1 & 80.87 & 56.07 & 66.72$\pm$1.42 \\
     & & PredictedClusterSampling & 0.1 & 82.11 & 58.43 & 69.78$\pm$2.08 \\
     \midrule
    \textbf{CLINC} & Standard & RandomSampling & 0.1 & 93.90 & 79.70 & 86.34$\pm$1.47 \\
     & & ClusterBased & 0.1 & 90.60 & 69.60 & 77.87$\pm$1.70 \\
     & & KnownClusterBased & 0.1 & 85.33 & 55.35 & 66.65$\pm$3.14 \\
     & & ClusterBasedSentenceEmb & 0.1 & 90.70 & 69.92 & 78.17$\pm$2.03 \\
     & & PredictedClusterSampling & 0.1 & 93.76 & 79.42 & 86.41$\pm$0.65 \\
     \cmidrule(r){2-7}
      & Sentence & RandomSampling & 0.1 & 93.80 & 79.06 & 85.76$\pm$1.17 \\
     & & ClusterBased & 0.1 & 90.25 & 67.23 & 74.25$\pm$1.83 \\
     & & KnownClusterBased & 0.1 & 84.95 & 53.34 & 63.89$\pm$1.55 \\
     & & ClusterBasedSentenceEmb & 0.1 & 90.03 & 66.23 & 73.60$\pm$1.73 \\
     & & PredictedClusterSampling & 0.1 & 93.60 & 78.57 & 85.43$\pm$0.96 \\
    \midrule
    \textbf{KidSpace} & Standard & RandomSampling & 0.2 & 71.40 & 48.26 & 58.55$\pm$4.22 \\
     & & ClusterBased & 0.2 & 68.13 & 39.26 & 53.48$\pm$4.47 \\
     & & KnownClusterBased & 0.2 & - & - & - \\
     & & KnownClusterBased & 0.4 & 69.76 & 55.52 & 61.10$\pm$6.15 \\
     & & ClusterBasedSentenceEmb& 0.2 & 66.92 & 38.46 & 53.87$\pm$6.58 \\
     & & PredictedClusterSampling & 0.2 & 70.53 & 45.33 & 56.80$\pm$4.52 \\
     \cmidrule(r){2-7}
      & Sentence & RandomSampling & 0.2 & 75.62 & 63.41 & 68.66$\pm$4.96 \\
     & & ClusterBased & 0.2 & 71.27 & 53.16 & 62.10$\pm$9.59 \\
     & & KnownClusterBased & 0.2 & 62.42 & 36.02 & 48.83$\pm$3.75 \\
     & & KnownClusterBased & 0.4 & 63.31 & 37.83 & 49.91$\pm$4.62 \\
     & & ClusterBasedSentenceEmb& 0.2 & 69.51 & 47.19 & 60.05$\pm$8.03 \\
     & & PredictedClusterSampling & 0.2 & 75.74 & 61.99 & 67.04$\pm$7.66 \\
     \cmidrule(r){2-7}
      & Sentence & RandomSampling & 0.1 & 65.21 & 38.04 & 48.85$\pm$2.47 \\
     & & ClusterBased & 0.1 & 62.60 & 35.41 & 49.62$\pm$5.40 \\
     & & KnownClusterBased & 0.1 & 63.32 & 38.00 & 50.95$\pm$4.65 \\
     & & ClusterBasedSentenceEmb & 0.1 & 61.96 & 34.98 & 47.76$\pm$2.85 \\
     & & PredictedClusterSampling & 0.1 & 66.98 & 40.55 & 51.20$\pm$4.15 \\
    \bottomrule
  \end{tabular}
  \caption{Semi-supervised DeepAlign Clustering Results with BERT Model, Data Balance/Augmentation and Seed Selection on BANKING, CLINC, and KidSpace datasets (averaged results over 10 runs with different seed values; known class ratio is 0.75 in all cases)}
  \label{bert-seed5-all}
\end{table*}

\appendix

\section{Appendix}

\subsection{Additional Experimental Results}
\label{appendix}

In addition to the Cluster-based Selection (CB) and Predicted Cluster Sampling (PCS) methods, we have explored other seed selection techniques compared with the Random Sampling. These are the Known Cluster-based Selection (KCB) and Cluster-based Sentence Embedding (CSE) methods. KCB is a variation of CB where we cluster into a number of known labels' subsets (based on known class ratio) and pick up certain \% of data (based on labeled ratio) from each cluster's data points. CSE, on the other hand, is another variation of CB where, instead of BERT word embeddings as the pre-trained representations, we use the sentence embeddings model before running K-Means (the rest is the same as the CB method).

Table~\ref{bert-seed5-all} presents detailed clustering performance results on three datasets using all five seed selection methods we explored, with varying labeled ratio and BERT embeddings (standard/BERT-base vs. sentence/SBERT models). In Table~\ref{ks-da-all}, we expand our analysis on the KidSpace dataset with data balancing/augmentation approaches on top of these five seed selection methods, once again with standard/sentence BERT embeddings. Table~\ref{bank-da-all} presents additional results on the BANKING dataset to compare data balancing/augmentation methods on top of standard vs. the sentence BERT representations.

\begin{table*}[!t]
  \centering
  \footnotesize
  \begin{tabular}{lccccccc}
    \toprule
    Dataset & BERT & Data Bal/Aug & Seed Selection & labeled\_ratio & NMI & ARI & ACC \\
    \midrule
    \textbf{KidSpace} & Standard & None & RandomSampling & 0.2 & 71.40 & 48.26 & 58.55 \\
     & & & ClusterBased & 0.2 & 68.13 & 39.26 & 53.48 \\
     & & & KnownClusterBased & 0.2 & - & - & - \\
     & & & ClusterBasedSentenceEmb& 0.2 & 66.92 & 38.46 & 53.87 \\
     & & & PredictedClusterSampling & 0.2 & 70.53 & 45.33 & 56.80 \\
     \cmidrule(r){2-8}
      & Standard & Paraphrasing & RandomSampling & 0.2 & 71.99 & 50.35 & 59.21 \\
     & & & ClusterBased & 0.2 & 68.04 & 39.80 & 55.06 \\
     & & & KnownClusterBased & 0.2 & 66.31 & 39.40 & 51.10 \\
     & & & ClusterBasedSentenceEmb& 0.2 & 67.44 & 39.49 & 54.42 \\
     & & & PredictedClusterSampling & 0.2 & 72.15 & 51.78 & 61.12 \\
     \cmidrule(r){2-8}
      & Standard & ParaMote & RandomSampling & 0.2 & 71.46 & 47.77 & 58.64 \\
     & & & ClusterBased & 0.2 & 67.82 & 39.59 & 54.56 \\
     & & & KnownClusterBased & 0.2 & 67.67 & 46.99 & 55.88 \\
     & & & ClusterBasedSentenceEmb& 0.2 & 66.64 & 39.61 & 53.82 \\
     & & & PredictedClusterSampling & 0.2 & 72.38 & 49.98 & 59.98 \\
     \cmidrule(r){2-8}
      & Sentence & None & RandomSampling & 0.2 & 75.62 & 63.41 & 68.66 \\
     & & & ClusterBased & 0.2 & 71.27 & 53.16 & 62.10 \\
     & & & KnownClusterBased & 0.2 & 62.42 & 36.02 & 48.83 \\
     & & & ClusterBasedSentenceEmb& 0.2 & 69.51 & 47.19 & 60.05 \\
     & & & PredictedClusterSampling & 0.2 & 75.74 & 61.99 & 67.04 \\
     \cmidrule(r){2-8}
      & Sentence & Paraphrasing & RandomSampling & 0.2 & 76.41 & 63.02 & 68.83 \\
     & & & ClusterBased & 0.2 & 70.71 & 48.19 & 60.88 \\
     & & & KnownClusterBased & 0.2 & 67.58 & 54.05 & 58.62 \\
     & & & ClusterBasedSentenceEmb& 0.2 & 70.93 & 52.60 & 62.67 \\
     & & & PredictedClusterSampling & 0.2 & 75.52 & 61.53 & 68.21 \\
     \cmidrule(r){2-8}
      & Sentence & ParaMote & RandomSampling & 0.2 & 76.28 & 61.20 & 68.09 \\
     & & & ClusterBased & 0.2 & 70.98 & 51.03 & 62.82 \\
     & & & KnownClusterBased & 0.2 & 67.13 & 49.47 & 56.64 \\
     & & & ClusterBasedSentenceEmb& 0.2 & 71.02 & 51.03 & 62.39 \\
     & & & PredictedClusterSampling & 0.2 & 76.33 & 62.05 & 68.21 \\
     \cmidrule(r){2-8}
     & Sentence & Aug (3x) & RandomSampling & 0.2 & 76.48 & 61.33 & 68.07 \\
     & & & PredictedClusterSampling & 0.2 & 76.37 & 58.97 & 68.78 \\
    \bottomrule
  \end{tabular}
  \caption{Semi-supervised DeepAlign Clustering Results with BERT Model, Data Balance/Augmentation and Seed Selection on KidSpace dataset (averaged results over 10 runs with different seed values; known class ratio is 0.75 in all cases)}
  \label{ks-da-all}
\end{table*}

\begin{table*}[!t]
  \centering
  \footnotesize
  \begin{tabular}{lccccccc}
    \toprule
    Dataset & BERT & Data Bal/Aug & Seed Selection & labeled\_ratio & NMI & ARI & ACC \\
    \midrule
    \textbf{BANKING} & Standard & None & RandomSampling & 0.1 & 79.22 & 52.96 & 63.84 \\
     & & & PredictedClusterSampling & 0.1 & 78.62 & 51.72 & 62.72 \\
     \cmidrule(r){2-8}
     & Standard & Paraphrasing & RandomSampling & 0.1 & 79.31 & 53.31 & 64.83 \\
     & & & PredictedClusterSampling & 0.1 & 78.79 & 52.41 & 64.62 \\
     \cmidrule(r){2-8}
     & Standard & ParaMote & RandomSampling & 0.1 & 79.62 & 54.08 & 65.37 \\
     & & & PredictedClusterSampling & 0.1 & 79.30 & 53.08 & 65.08 \\
     \cmidrule(r){2-8}
      & Sentence & None & RandomSampling & 0.1 & 82.96 & 60.72 & 71.27 \\
     & & & PredictedClusterSampling & 0.1 & 82.11 & 58.43 & 69.78 \\
     \cmidrule(r){2-8}
     & Sentence & Paraphrasing & RandomSampling & 0.1 &	83.00 & 60.95 & 71.95 \\
     & & & PredictedClusterSampling & 0.1 & 82.20 & 58.86 & 69.62 \\
     \cmidrule(r){2-8}
     & Sentence & ParaMote & RandomSampling & 0.1 & 82.58 & 59.54 & 69.92 \\
     & & & PredictedClusterSampling & 0.1 & 81.88 & 58.13 & 69.74 \\
     \cmidrule(r){2-8}
     & Sentence & Aug (3x) & RandomSampling & 0.1 & 82.94 & 60.78 & 71.66 \\
     & & & PredictedClusterSampling & 0.1 & 81.69 & 58.18 & 69.99 \\
    \bottomrule
  \end{tabular}
  \caption{Semi-supervised DeepAlign Clustering Results with BERT Model, Data Balance/Augmentation and Seed Selection on BANKING dataset (averaged results over 10 runs with different seed values; known class ratio is 0.75 in all cases)}
  \label{bank-da-all}
\end{table*}

\end{document}